\documentclass{article}
\usepackage{spconf,amsmath,graphicx}
\usepackage{booktabs}

\title{BridgeNets: Student-Teacher Transfer Learning based on
       Recursive Neural Networks and its Application to Distant Speech Recognition}
\name{Jaeyoung Kim, Mostafa El-Khamy, Jungwon Lee}
\address{Samsung Semiconductor, Inc. USA\\
	Emails:\{jaey1.kim, mostafa.e, jungwon2.lee\}@samsung.com}

\begin{document}
%
\maketitle

\begin{abstract}
Despite the remarkable progress achieved on automatic speech recognition,
recognizing far-field speeches mixed with various noise sources
is still a challenging task.
In this paper, we introduce novel student-teacher transfer learning, BridgeNet which
can provide a solution to improve distant speech recognition. There are two key features
in BridgeNet.
First, BridgeNet extends traditional student-teacher frameworks by providing multiple hints from
a teacher network. Hints are not limited to the soft labels from a teacher network.
Teacher's intermediate feature representations can better guide a student network
to learn how to denoise or dereverberate noisy input.
Second, the proposed recursive architecture in the BridgeNet can iteratively
improve denoising and recognition performance.
The experimental results of BridgeNet showed significant improvements in tackling the distant
speech recognition problem, where it achieved up to 13.24\% relative WER reductions on AMI corpus
compared to a baseline neural network without teacher's hints.
\end{abstract}


\section{Introduction}
\label{sec:intro}
Distant speech recognition (DSR) is to recognize human speeches in the presence of noise,
reverberation and interference caused mainly by the large distance between speakers and
microphones.
DSR is a challenging task especially due to unavoidable mismatches in signal quality between
normal close-talking and far-field speech signals. Traditional speech
recognizers trained with speech samples from close-talking microphones show
significant performance drops in recognizing far-field signals.

There have been great efforts to improve DSR performance. Traditional front-end approaches interconnect
multiple independent components such as speech
enhancer~\cite{brandstein2013microphone,makino2007blind},
acoustic speech detector~\cite{sohn1999statistical,ramirez2004efficient},
speaker identification~\cite{reynolds2000speaker,garcia2011analysis}
and many other blocks before a speech recognition module. 
The interconnected
components denoise and dereverberate far-field speeches to generate enhanced data.
A major issue in these approaches is the mismatch between combined components
because they are independently optimized without consideration of each other.

Many end-to-end methods are proposed to overcome the issue of front-end approaches
by jointly optimizing multiple components in the unified framework. Among them,
we discuss two popular approaches relevant to our method.

Multi-task denoising
\cite{qian2016investigation,ravanelli2016batch,ravanelli2017network}
jointly optimizes denoising and
recognition sub-networks using synchronized clean data.
It minimizes the weighted sum of two loss functions:
cross-entropy loss from recognition sub-network output and
mean square error (MSE) loss between denoising sub-network output and
clean data. Although multi-task denoising showed some improvements on DNN acoustic models,
minimizing MSE between raw acoustic data and high-level abstracted features is
often unsuccessful. Its performance depends heavily on the underlying acoustic models. 

Knowledge distillation (KD) ~\cite{hinton2015distilling,romero2014fitnets} transfers
the generalization ability of a bigger teacher network to a typically much smaller
student network. It provides soft-target information computed by the teacher network,
in addition to its hard-targets, so the student network can learn to generalize similarly.
Generalized distillation (GD)~\cite{lopez2015unifying,markov2016robust_1,markov2016robust_2}
extends distillation methods by training a teacher network with separate clean data.
A student network is trained on noisy data and, at the same time, guided
by the soft-labels from a teacher which has access to synchronized clean speech.
The generalized distillation methods showed decent performance on CHiME4 and Aurora2 corpora.

In this paper, we propose novel student and teacher
transfer learning, BridgeNet which further extends knowledge distillation~\cite{hinton2015distilling}.
There are two key features in BridgeNet.
\begin{itemize}
\item
BridgeNet provides multiple hints from a teacher network. KD and GD methods
utilize only teacher's soft labels. BridgeNet provides teacher's
intermediate feature representations as additional hints, which can properly regularize a student
network to learn signal denoising.
\item
The proposed recursive architecture in the BridgeNet can iteratively refine recognition and
denoising performance. As ASR performance can be enhanced by signal denoising,
signal denoising can be also improved by reference to ASR output.
The proposed recursive
architecture enables bi-directional information flows between signal denoising and 
speech recognition functions by simple network cascading.
\end{itemize}

The experimental results confirm the effectiveness of BridgeNet by showing that
BridgeNet with multiple hints presented up to 10.88\% accuracy improvements
on the distant speech AMI corpus. With a recursive architecture, BridgeNet
achieved up to 13.24\% improvements.



\section{BridgeNets}
\label{sec:bridgenet}

\subsection{Network Description}
\begin{figure}[h]
  \centering
  \centerline{\includegraphics[width=7.0cm]{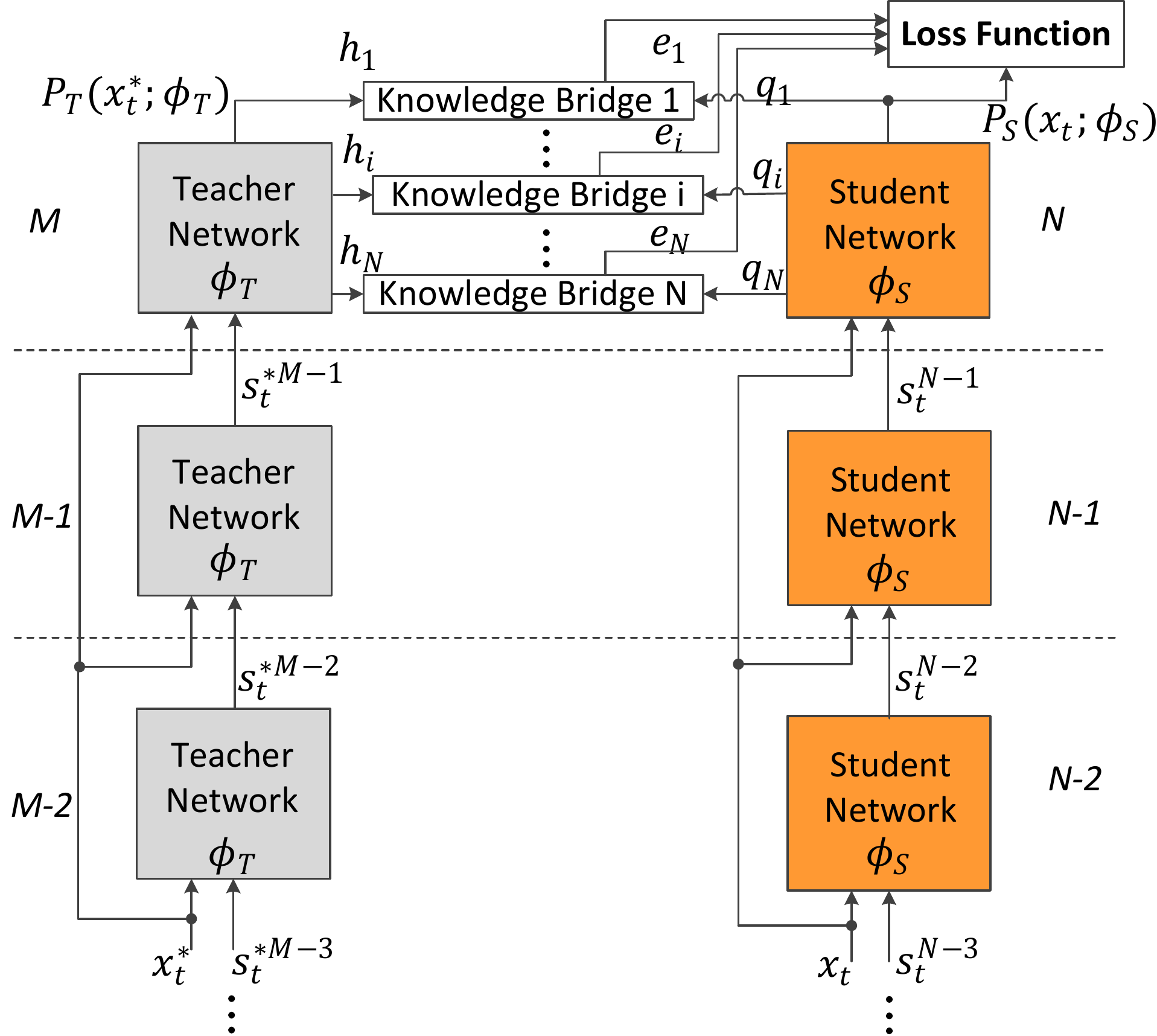}}
\caption{Conceptual Diagram of BridgeNet}
\label{fig:br_overview}
\end{figure}

BridgeNet provides novel student-teacher transfer learning based on a new recursive
architecture to deploy the learning-from-hints paradigm~\cite{abu1993hints}.
Figure~\ref{fig:br_overview} presents a high-level block diagram of BridgeNet.
Both student and teacher networks are constructed from a recursive network.
They don't need to have the same recursion number.
Typically, a teacher network can have more recursions because its complexity only matters
during training stage.

BridgeNet uses a collection of triplets as training data: $\left(x_t^*,x_t,y_t \right)$.
$x_t^*$ is enhanced or less noisy data, $x_t$ and $y_t$ are noisy data and their labels.
A teacher network is trained with $x_t^*$ and $y_t$ pairs. 
The trained teacher network provides its internal feature representations
as hints to a student network. Knowledge bridges are connections between
teacher's hints and student's guided layers. The connected two layers at the knowledge bridges
should have similar level of abstraction.
For example,
the student's knowledge bridge of LSTM3 in Figure~\ref{fig:cnn_lstm} should be connected to the similar
LSTM output at the teacher network.

An error measure $e_i$ of how a feature representation $q_i$ from a student network agrees
with the hint $h_i$ is computed at the knowledge bridge as a MSE loss,
\begin{equation}
e_{i}(\phi_S) = \sum_{t=1}^{L}\left\|h_{i}(x^{*}_{t})-q_{i}(x_{t};\phi_S)\right\|^{2}
\label{eq:kb_1}
\end{equation}
where $\phi_S$ is the learnable parameters of a student network.
Since $h_1$ and $q_1$ are softmax probabilities of teacher and student networks,
the cross-entropy loss is used for $e_1$ instead.
\begin{equation}
e_1(\phi_S) = -\sum_{t=1}^{L}\left(P_{T}\left(x_{t}^{*};\phi_T\right)\right)^{T}\log P_{S}(x_{t};\phi_S)
\end{equation}
The parameters of the student network are then optimized by minimizing a weighted
sum of all corresponding loss functions,
\begin{equation}
L(\phi_S) = \sum_{i=1}^{N}\alpha_{i} e_{i}(\phi_S)
\end{equation}
where $\alpha_i$ is a predetermined weighting factor for $e_{i}$.

Since student and teacher networks have multiple recursions, the same knowledge bridges can be
repeatedly connected for every recursion.
However, any knowledge bridge added at the intermediate recursion always degraded performance.
BridgeNet adds knowledge bridges only at the last recursion as shown in Figure~\ref{fig:br_overview}.


\subsection{Recursive Architecture}
\label{sec:recursive}

In this section, we present a new recursive architecture.
A recursive neural network is popularly used in sentence parsing, sentimental analysis,
sentence paraphrase and many other areas. It applies the same set of weights
recursively over a structure. Its concept is similar to a recurrent network but there is a clear
difference in that a recursive neural network can traverse a given structure in any topological order.

\begin{figure}[htb]
\begin{minipage}[b]{0.32\linewidth}
  \centering
  \centerline{\includegraphics[width=1.7cm]{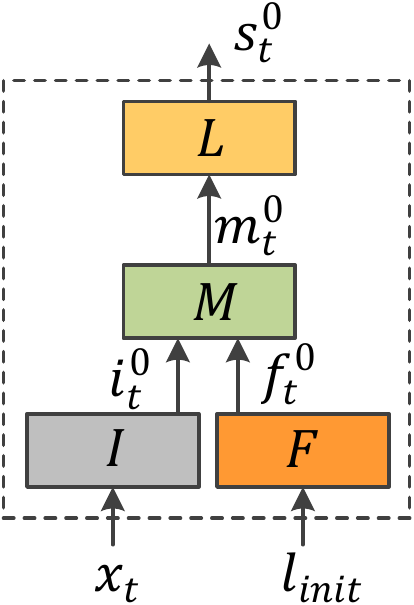}}
  \centerline{(a) $R=0$ }\medskip
\end{minipage}
\hfill
\begin{minipage}[b]{0.32\linewidth}
  \centering
  \centerline{\includegraphics[width=2.0cm]{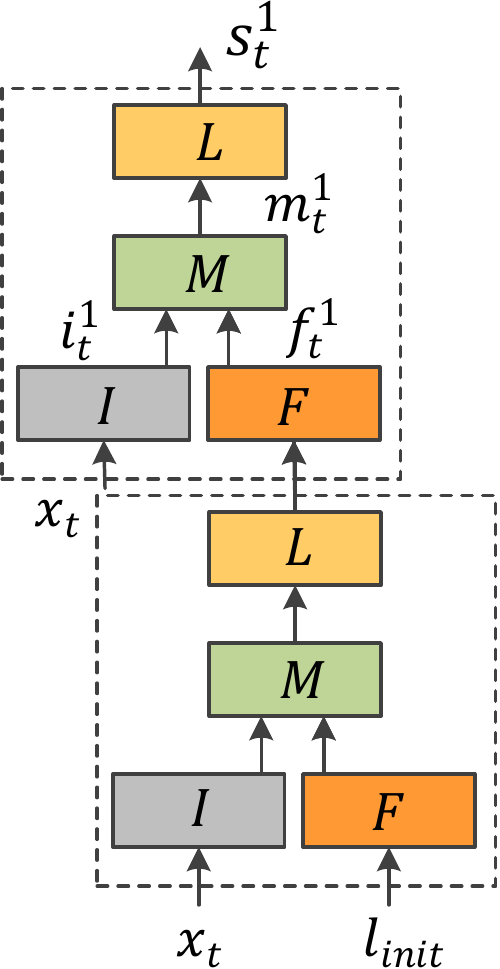}}
  \centerline{(b) $R=1$}\medskip
\end{minipage}
\hfill
\begin{minipage}[b]{0.32\linewidth}
  \centering
  \centerline{\includegraphics[width=2.3cm]{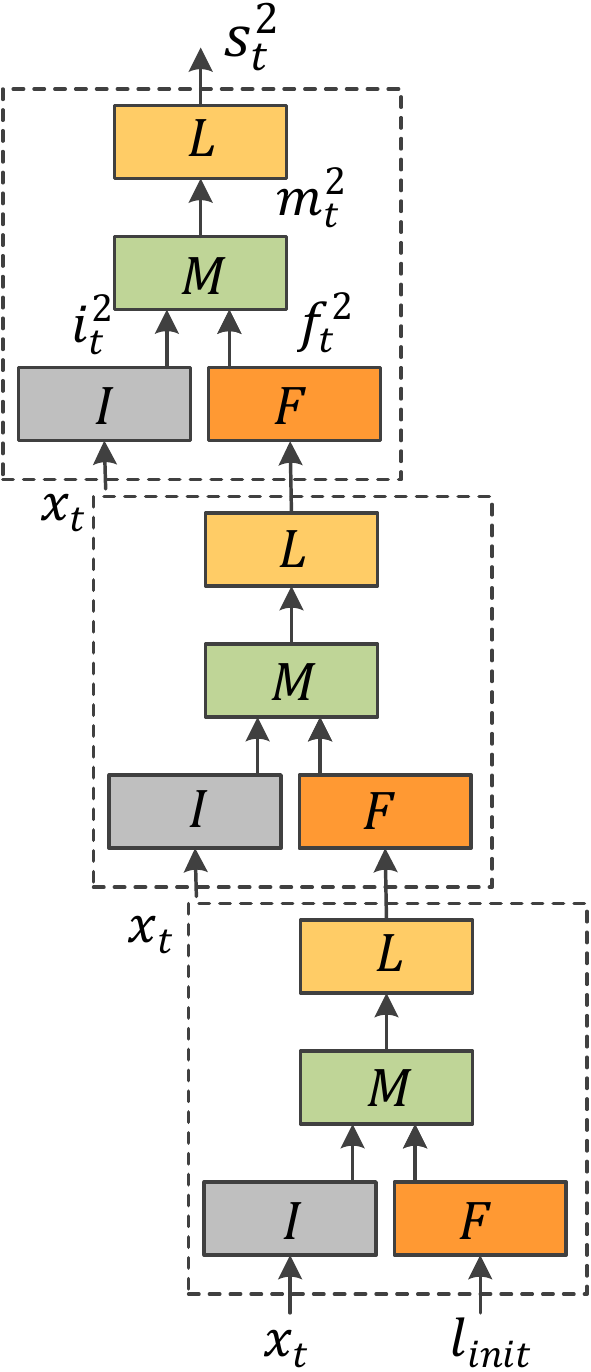}}
  \centerline{(c) $R=2$}\medskip
\end{minipage}
\caption{ Unrolling of a Recursive Network: $R$ is the number of recursions.
(a), (b) and (c) show how a recursive network is unrolled in the depth direction.
The blocks with the same color share the same weights.}
\label{fig:recursive_a}
\end{figure}

Figure~\ref{fig:recursive_a} (a) shows building blocks of a proposed recursive architecture.
It is composed of four sub-blocks: $I$ and $F$ take acoustic features and feedback states as their
input, $M$ merges $I$ and $F$ outputs and $L$ produces recognized phone states.
Each block can be any type of network. $i_t^n$, $f_t^n$, $m_t^n$ and $s_t^n$ represents output
for the corresponding sub-blocks. $n$ indicates the recursion number.
$l_{init}$ is a zero vector used as input for the zero recursion. 

The advantage of this sub-block division enables a network to recurse with heterogeneous input
and output types. For example, a typical acoustic model has context-dependent phones as a
network output. This output cannot be fed into an input for the next recursion because the network
input is an acoustic signal that is totally different from phone states. The proposed architecture
provides two different input paths. They are processed independently and merged later at the $M$. 

Figure~\ref{fig:recursive_a} presents how to unroll the proposed recursive network in the depth
direction. $R$ implies the number of recursion. The same input $x_t$ is
applied to the network for each recursion. This repeated input acts as a global shortcut path
that is critical to train a deep architecture. Our proposed recursive network can be formulated
as follows:
\begin{equation}
\label{eq:recursive_0}
m^n_t = g\left( W_{1}\cdot i^{n}_t(x_t) + W_{2}\cdot f_{t}^{n}\left( s_{t}^{n-1}\right) + b\right)
\end{equation}
$W_{1}$, $W_{2}$ and $b$ are the internal parameters of $M$. Two paths are affine-transformed and
added together before going into nonlinear function $g$. Compared with the recursive residual
network proposed in~\cite{taiimage}, our model has two differences. First, the model in~\cite{taiimage}
can only recurse with homogeneous input and output. Second, a global shortcut path is always added
with the output of the prior recursion in~\cite{taiimage} but our model allows to flexibly
combine two heterogeneous inputs.
Simple addition is a special case of Eq.~\ref{eq:recursive_0}.

\begin{figure}[h]
  \centering
  \centerline{\includegraphics[width=5.3cm]{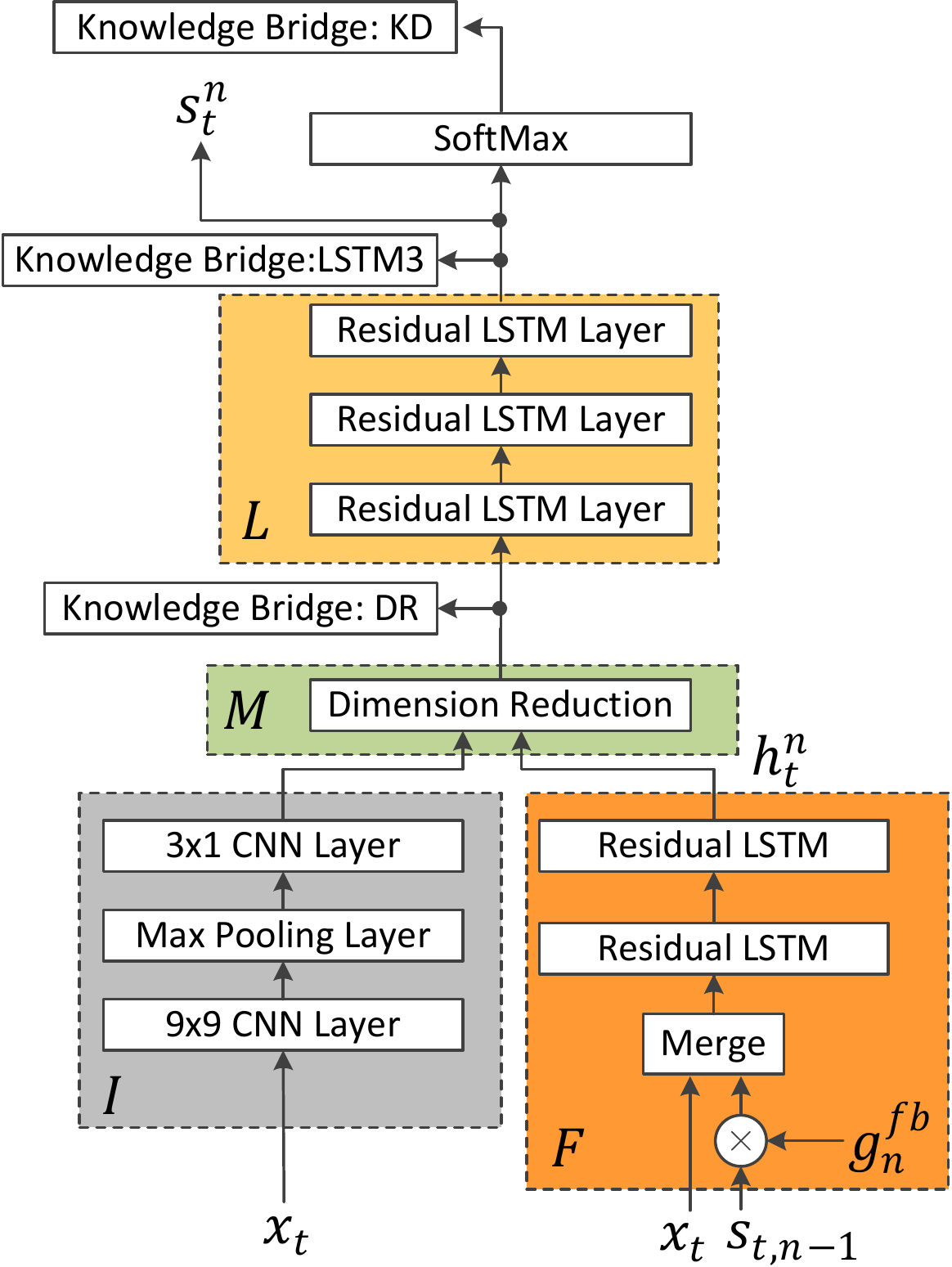}}
\caption{CNN-LSTM recursive network: knowledge bridges are added at the last recursion.}
\label{fig:cnn_lstm}
\end{figure}

Figure~\ref{fig:cnn_lstm} shows how the Bridgenet concept is applied to the recursive network of
Figure~\ref{fig:recursive_a}.
It has four components:
CNN layers ($I$), first LSTM layers ($F$), second LSTM layers ($L$) and
dimension reduction layer ($M$).
Since feedback phone states and acoustic input don't have
correlations in frequency and time directions, they cannot be fed into
the same CNN layers. Instead, feedback phones are separately processed in $F$ controlled
by a gate network, $g_n^{fb}$. Its formulation is referred from~\cite{chung2015gated},
\begin{equation}
g^{fb}_{n} = \sigma\left( w_x x_t + w_s s_{t}^{n-1} + w_{h} h_{t-1}^n \right)
\end{equation}
where $s_{t}^{n-1}$ is a feedback state from the $(n-1)^{th}$ recursion, $h_{t}^n$ is
an the output of $F$ at the $n^{th}$ recursion and $w_x$, $w_s$ and $w_h$ are weights to be learned.
Two input paths are combined
later at the dimension reduction layer. The dimension reduction layer is a fully-connected
one to merge them and reduce their dimensions for the second LSTM block, $L$.

A residual LSTM~\cite{kim2017residual} is used for $F$ and $L$ sub-blocks.
It has a shortcut path between layers to avoid vanishing or exploding gradients commonly
happening to deep networks. It was shown that residual LSTM outperforms plain LSTM
for deep networks.

\section{Experiments}
\subsection{Experimental Setup}

AMI corpus~\cite{carletta2005ami} provides 100 hours meeting conversations
recorded both by individual headset microphones (IHM) and single distant
microphones (SDM). IHM data is cleanly recorded but SDM has high noise
and other speaker's interferences. SDM can be improved by beamforming multiple
SDM channels, which becomes MDM data.
Since IHM, SDM and MDM corpora are synchronously recorded,
an alignment label generated by one corpus type can be used to train a network
with any other corpus. BridgeNet is trained with a clean alignment from IHM.

Kaldi~\cite{Povey_ASRU2011} and
Microsoft Cognitive Toolkit (CNTK)~\cite{yu2014introduction} are used to train and decode
BridgeNet.
Log filterbank amplitudes with 80 dimensions are generated as feature vectors.
They are stacked as 9 frames to be fed into BridgeNet.
Residual LSTMs in BridgeNet has 1024 memory cells and 512 hidden nodes.
The final softmax output has 3902 context-dependent phone classes. Two CNN layers
has 9x9 and 3x1 kernels with 256 feature maps, respectively.

Since SDM or MDM corpus is a meeting conversation between multiple speakers,
we provide two types of word error rates (WER): all-speakers and main-speaker WERs.
The all-speakers WER is to decode up to 4 concurrent speeches, which
is a big challenge considering training procedure only focuses on a main
speaker. The main-speaker WER is to decode single main speaker at each time frame, which is
more realistic performance measure for SDM or MDM corpus. We provides both WERs
in the later evaluations.

\subsection{BridgeNet and Multi-Task Denoising on AMI}
\label{sec:multi}
Table\ref{tab:mtd} provides WER evaluation of multi-task denoising on SDM corpus.
Multi-task denoising shows only 0.7\% and 0.6\% main-speaker WER reduction on DNN and
CNN-LSTM, which is contrary to the bigger improvement observed in~\cite{qian2016investigation}.
The trained DNN has 8 layers and each layer has 2048 neurons except
bottleneck layers, which is the same as the model in~\cite{qian2016investigation}.
The main difference is that our DNN model showed significantly lower WERs. It is conjectured
that the gain from multi-task denoising decreases for better acoustic model.
Next, CNN-LSTM is trained with a clean alignment from IHM corpus. The main-speaker WER
of CNN-LSTM got improved more
than 9\% simply changing alignment labels. However, multi-task denoising on the
improved CNN-LSTM degraded WER from 37.8\% to 38.2\%.
\begin{table}
\caption{Multi-Task Denoising on SDM: $\text{CNN-LSTM}^{*}$ was
trained with a clean alignment from IHM. Other models used
a noisy alignment from SDM}
\label{tab:mtd}
\vspace{0.2cm}
\centering
\begin{tabular}{l r r }
\toprule
 Acoustic Model   &  WER (all) & WER (main) \\
 \midrule
 DNN   & 59.1\% & 50.5\% \\
 DNN, denoised   & 58.7\% & 50.2\% \\
 CNN-LSTM  & 50.4\% & 41.6\% \\
 CNN-LSTM, denoised & 50.1\% & 41.4\%\\
 \midrule
$\text{CNN-LSTM}^{*}$ & 46.5\% & 37.7\%\\
$\text{CNN-LSTM}^{*}$, denoised& 46.9\% & 38.2\%\\
\bottomrule
\end{tabular}
\end{table}

Table~\ref{tab:br_sdm} presented BridgeNet WER results on SDM corpus.
A CNN-LSTM is a baseline network.
KD, DR and LSTM3 are knowledge bridges shown in Figure~\ref{fig:cnn_lstm}.
KD in Table~\ref{tab:br_sdm} means a BridgeNet with only knowledge distillation connection.
Likewise, KD+DR and KD+DR+LSTM3 imply BridgeNets with corresponding knowledge bridges.
BridgeNets with R0 have no recursion both for student and teacher networks.
For BridgeNets
with R1, a student network has one recursion but a teacher network has two recursions, where
knowledge bridges were only added to the last recursion.

BridgeNet on the non-recursive network, R0, with KD, DR and LSTM3 showed 6.9\% and 1.6\% relative WER
reduction over CNN-LSTM and BridgeNet with KD, respectively. These results showed
Knowledge bridges at the intermediate layers further improved a student network
by guiding student's feature representations. For R1,
KD+DR+LSTM3 showed 3.7\% relative WER reduction over KD+DR+LSTM3 without recursion.
Compared with CNN-LSTM and KD, KD+DR+LSTM3 with R1 provided
10.34\% and 5.04\% improvements, respectively. The recursive architecture
significantly boosted the performance of a student network.  

Table~\ref{tab:br_mdm} presented BridgeNet WER results on MDM data. MDM data was
formed by beamforming 8 channel SDM data using BeamformIt~\cite{anguera2007acoustic}.
A student network is trained with beamformed MDM training data and also evaluated
with beamformed evaluation data.
Similar to the SDM results, BridgeNet provides significant improvements.
For R0, KD+DR+LSTM3 showed 5.29\% and 2.72\% relative WER reduction over CNN-LSTM and KD.
With recursions, its gain increased as 13.24\% and 10.88\% compared with the baseline
and KD.
\begin{table}
\caption{BridgeNet: A teacher network is trained with IHM data and
a student network is trained with SDM data. Rn means the network has n recursions.
(e.g. baseline CNN-LSTM with R2 has two recursions)}
\label{tab:br_sdm}
\vspace{0.2cm}
\centering
\begin{tabular}{l r r }
\toprule
 Acoustic Model   &  WER (all) & WER (main) \\
 \midrule
 CNN-LSTM (baseline), R0  & 46.5\% & 37.7\% \\
 KD, R0   & 44.8\% & 35.7\% \\
 KD+DR, R0  & 44.1\% & 35.3\% \\
 KD+DR+LSTM3, R0 & 44.0\% & 35.1\% \\
 \midrule
 CNN-LSTM, R2  & 45.8\% & 36.9\% \\
 KD, R1   & 43.7\% & 34.7\% \\
 KD+DR, R1  & 43.4\% & 34.7\% \\
 KD+DR+LSTM3, R1 & 42.6\% & 33.8\% \\
\bottomrule
\end{tabular}
\end{table}
\begin{table}
\caption{BridgeNet: A teacher network is trained with clean IHM data and
a student network is trained with MDM data.}
\label{tab:br_mdm}
\vspace{0.2cm}
\centering
\begin{tabular}{l r r }
\toprule
 Acoustic Model   &  WER (all) & WER (main) \\
 \midrule
 CNN-LSTM (Baseline), R0   & 43.4\% & 34.0\% \\
 KD, R0   & 42.8\% & 33.1\% \\
 KD+DR, R0  & 42.3\% & 32.5\% \\
 KD+DR+LSTM3, R0 & 41.8\% & 32.2\% \\
 \midrule
 CNN-LSTM, R2  & 43.0\% & 33.3\% \\
 KD, R1   & 40.4\% & 30.8\% \\
 KD+DR, R1  & 39.5\% & 29.9\% \\
 KD+DR+LSTM3, R1 & 39.3\% & 29.5\% \\
\bottomrule
\end{tabular}
\end{table}

\section{Conclusion}

This paper proposes a novel student-teacher transfer learning, BridgeNet. BridgeNet
introduces knowledge bridges that can provide 
a student network with enhanced feature representations
at different abstraction levels. BridgeNet is also based on
the proposed recursive architecture, which enables to iteratively improve signal denoising
and recognition.
The experimental results confirmed training with
multiple knowledge bridges and recursive architectures significantly improved distant
speech recognition.
\bibliographystyle{IEEEbib}
\bibliography{strings,refs}

\end{document}